# CNN inference acceleration using dictionary of centroids.

**D.Babin, I.Mazurenko, D.Parkhomenko, A.Voloshko.**
Faculty of Mechanics and Mathematics, Moscow State University,
denis.parkhomenko@intsys.msu.ru
ivan@mazuernko.ru

**Abstract.**
It is well known that multiplication operations in convolutional layers of common CNNs consume a lot of time during inference stage. In this article we present a flexible method to decrease both computational complexity of convolutional layers in inference as well as amount of space to store them. The method is based on centroid filter quantization and outperforms approaches based on tensor decomposition by a large margin. We performed comparative analysis of the proposed method and series of CP tensor decomposition on ImageNet benchmark and found that our method provide almost 2.9 times better computational gain. Despite the simplicity of our method it cannot be applied directly in inference stage in modern frameworks, but could be useful for cases calculation flow could be changed, e.g. for CNN-chip designers.

## 1. Introduction.

During last several years we are observe a fast CNN evolution. Neural networks become extremely popular. They cover more and more new areas while going deeper in already existing ones: classification, detection, re-identification etc. Despite several successful complexity reduction methods described in literature, usually the scaling problem is solved by enlarging amount of computational resources (most common number of GPUs). We hope our work will help engineers to make a step aside from this practice and push them to towards more efficient computational power usage. Our experiments show that VGG-16 network inference can be improved by a factor of 10.9 in terms of MAC operations without only a 1% of accuracy loss. All tests were performed on ImageNet benchmark, codes we used will be posted in open access.

Several attempts to decrease complexity of the networks are known in literature. All approaches can be roughly split in into following categories: low-rank decomposition of the CNN layers [1], weights binarization and quantization [2], pruning and sharing [3]. Each category has it's own values (e.g. pruning in parallel address over-fitting problem) and drawbacks, but all of them designed to reduce only the size required to store the networks. Among methods designed to decrease the calculational complexity of networks we can outline following categories: low-rank decomposition of the CNN layers [1], weights binarization and quantization [2], pruning and sharing [3], knowledge Distillation [4,5] and mutual learning [6]. Our approach is complementary and can be used in combination with any of them. The method we propose shows good performance in both computational performance and size complexity, which is reduced more than 15 times. However, the method requires changes in convolution calculation flow. We expect it would be helpful for designers of CNN-related hardware or for researchers, who can vary procedure of convolution computation.

The rest of the paper is organized as follows. Section 2 shortly reviews existing methods to address issues of the computational complexity and compression of the convolutional neural networks. Section 3 describes our method in details. Comparative analysis and experiment results are given in Section 4. Section 5 concludes the paper. To validate our methodology, we provide access to codes with comparative analysis of our method and tensor-factorization based approach.

## 2. Section 2.

One of the most popular method to shrink big amount of data is factorization. There is a bunch of methods on applying decomposition techniques on convolution kernels [1,6,10,12]. The convolution kernel can be represented as a 4D tensor. Ideas based on tensor decomposition are derived by intuition that there is a big amount of redundancy in a large 4D tensor. Fully connected layers can be considered as a 2D matrices, and low-rankness can also help. The variety of factorization methods is motivated by their often success. Most of research is done in the compression domain. I.e. factorization is used to save storage used for tensor/neural networks [1,3,12]. Various tensor decompositions are also applied for the networks inference acceleration [6,10,11]. The idea is that convolutions can be calculated with fewer number of operations due to special structure of decomposed tensors. We will describe this better in Section 4. Depending on the network architecture, decomposition type and other parameters, one is able to speed-up computation in several times and compress the network in dozens of times. Main drawbacks of this approaches are: low-rank decomposition is a computationally expensive operation, no theoretically proven way to get the best (in terms of resulting accuracy) decomposition, most tensor-related problems are NP-hard [8].

Different authors use different approaches to evaluate acceleration/compression gain, which complicates comparison. That is why we have to implement the tensor decomposition by our means. We choose tensor decomposition as a reference method to compare performance.

Another bunch of methods is known as «pruning and sharing». Methods of this family address the over-fitting issue and the network compression. The idea is to prune redundant, non-informative weights in a pre-trained CNN model. E.g. [13] explored the redundancy among neurons, which is removed using data-free pruning method. Another work [14] proposes to use low-cost hash function for weights grouping into buckets for parameter sharing. Chen et al. [14] also check this approach on VGG-16 and succeed to compress VGG in 49 times. Our method shows 10.9 compression ratio together with acceleration inference time by a factor of 10.9 as well. There is a also growing interest in training compact CNNs with sparsity constraints, which are introduced as regularization rules [15]. To the minuses of this approaches can be attributed: all pruning criteria require manual setup of sensitivity for layers. Methods with sparsity constraints require network re-training and usually larger amount of iterations to converge.

The next popular bunch of methods is based on quantization and binarization. The network quantization compresses the original network by reducing number of bits required to represent each weight. Vanhoucke et al [16] had shown that 8-bit quantization of parameters can result in significant speed-up with minimal loss of accuracy. The work [17] uses 16-bit fixed-point representation in CNN training, which reduced memory usage and floating point operations with little loss of accuracy. Binarization can be considered as an extreme case of 1-bit quantization [2]. One should notice, that mathematically, quantization can be considered as a special case of low-rank filter decomposition. Not surprisingly, that factorization shows better results. Quantization for the inference acceleration requires special hardware support (e.g. AVX,SSE) or changing computational primitives.

Another interesting approach for neural network acceleration and compression is the knowledge distillation (KD). Originally it was proposed by Caruana et al. [4], and further supported by researchers community [5,18,19]. Caruana et al. [4] trained an ensemble of strong classifiers and reproduce the output of a large original network. The idea of KD is to compress a large and deep neural networks into shallower ones, which reproduce the function learned by the complex initial model. This direction is deep and orthogonal to the method that we use in our work. That is why we prefer not to go deeper in details. The reader who wants to get more familiar with the KD can search for more details in [4,5,18,19,20]. Usually, distilled model assumptions are too hard to show the state-of-the-art performance.

To our knowledge, the approach that we propose in current article, shows one of the best performance in terms of complexity reduction, neural network acceleration and resulting model accuracy. The main drawback of this method is that it can not be applied directly in modern deep learning frameworks. The method requires a special convolution implementation, which could be done both by software or hardware means. We expect that this method should be interesting for CNN-related hardware designers.

## 3. Section 3.

We propose to apply a filter quantization followed by clustering in a special way to form a dictionary, which will be used for convolution computations in score estimation. Consider convolutional layer of some neural network, and related convolutional kernel. Let the kernel $W=\{\omega_{ijkl}\}$ be of a size $[k_1,k_2,k_3,k_4]$, $0 \leq i \leq k_1$, $0 \leq j \leq k_2$, $0 \leq k \leq k_3$, $0 \leq l \leq k_4$, and input layer size is $[L_1,L_2,L_3,L_4]$. Let us also consider an integer $M>1$, kernel $\widehat{W}$ of size $[k_1,k_2,k_3,k_4]$ and dictionary $D = \{d_i \mid d_i = (d_i[1],..,d_i[M])\}$ of finite size with following properties:
1) $\widehat{W}$ consists only of $D$ elements :
    for every $i,j,k,l$ there exists a $d \in D$, that $\omega_{ijkl}$ which belongs to $D$
2) $\widehat{W}$ can be splitted into disjoint combination of dictionary elements
3) Every element of the $D$ is present in $\widehat{W}$.

If we find such a dictionary $D$ and $\widehat{W}$, that the Frobenius norm of the tensor subtractions $|W - \widehat{W}|_F$ is as small as possible, then we will approximate filter $W$ by filter $\widehat{W}$, which has deterministic structure given by dictionary. As one can see, a solution always exists. If we draw $W$ as a vector and split it into parts of size $M$, they will form a dictionary $D$, which gives a zero error in approximation. However, the size of such a dictionary is too large.

Experiments show that modern CNNs has large redundancy inside convolutional kernels in almost every layer. Usage of this intra-layer redundancy allows to decrease dictionary size. As a consequence, number of multiply-accumulate (MAC) operations required for convolution reduces, thus compressing network in terms of stored data amount. Below, we estimate acceleration and compression impact of our method under assumption that convolutional strides in every layer of the network are equal to 1. For VGG-16 this condition is satisfied.

To search $D$ and $\widehat{W}$ we use k-means clustering approach. We found that exploiting of natural redundancy of the visual images tends to better decomposition, i.e. dictionary of smaller size. RGB components of the image have high pixel correlation, which propagates to next layers of the network.

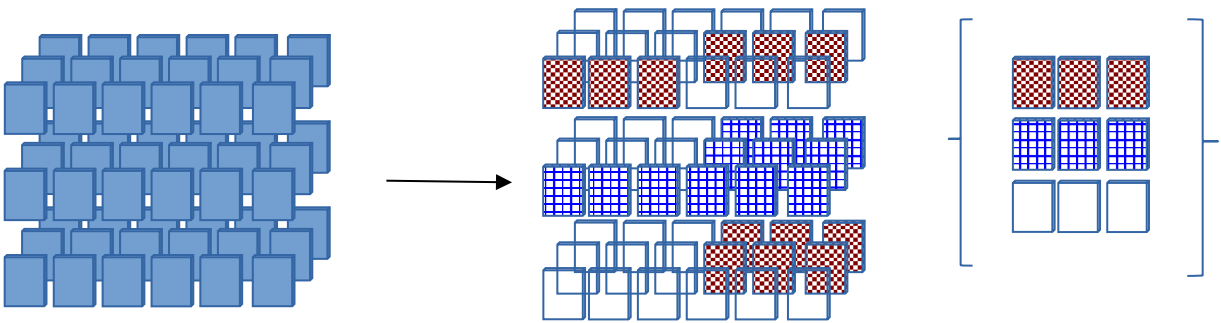

Fig.1. Structure of the original and decomposed kernels of size (3,3,6). Dictionary D consist of 3 vectors depicted by various hatching at the right.

Consequently, we split the kernel $W$ along the axis parallel to data propagation in the network and perform k-means clustering to find a predefined number of parts to form dictionary $D$.

$\widehat{W}$ reconstruction is performed by a choosing closest centroid. i.e. to reconstruct $(\widehat{W}_{i,j,k,l}, \ldots, \widehat{W}_{i+M-1,j,k,l})$ we use a centroid $d \in D$, which is the closest to $(W_{i,j,k,l}, \ldots, W_{i+M-1,j,k,l})$ in terms of Euclidian norm. In other words:
$(\widehat{W}_{i,j,k,l}, \ldots, \widehat{W}_{i+M-1,j,k,l}) = argmin_{d \in D}( d-(W_{i,j,k,l}, \ldots, W_{i+M-1,j,k,l}) )$

Below we estimate amount of memory, which is required for kernel storage in approximation of $\widehat{W}$ with help of dictionary *D* and overall memory gain. Consider that values of kernel has 32 bit representation. At this moment kernel approximation can be given by dictionary *D* and tensor *O* of size $([k_1/M]+1,k_2,k_3,k_4)$, which maps dictionary elements into kernel $\widehat{W}$. Elements of the order tensor *O* can be represented by $[log2(|D|)]+1$ bits, so number of bits required to store $\widehat{W}$, *O* and *D* can be estimated as: $32*M*|D|+([log2(|D|)]+1)*([k_1/M]+1)*k_2*k_3*k_4$. To store tensor $\widehat{W}$ in common form required: $32*k_1*k_2*k_3*k_4$

To use the described method, the convolution operation has to be changed. It can be done both in software or hardware. If speak about hardware, the main chip complexity measure is the gates count. All operations in hardware are implemented using gates. More complex operations (e.g. multiplication) requires more gates to be implemented.

In order to use less multiply-accumulate operations, we first compute convolution of the input layer with every element of the dictionary *D*, after we accumulate result according to map tensor *O*. Figure 2 shows this process. The proposed method of convolution computation requires additional memory overhead to store intermediate convolutions with dictionary elements. Still, in terms of gates count it is more attractive than common convolution computation because number of elements in dictionary is usually small.

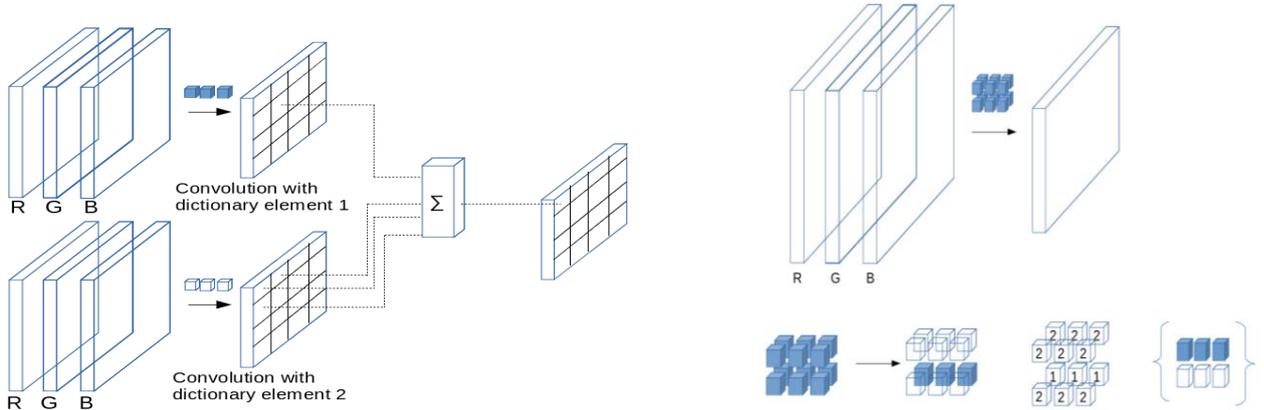

Fig. 2. Convolution calculation using dictionary *D* (left) and traditional method of convolving input RGB image with filter of size 2x2x3 (right). Dictionary of size 2 is depicted

Next, we estimate number of MAC operations required to compute convolution layer output in common way and using dictionary D and map tensor O. Consider kernel of size $[k_1,k_2,k_3,k_4]$, input layer size $[L_1,L_2,L_3,L_4]$ and dictionary element of length M. Convolution computation requires $L_1*L_2*L_3*L_4*k_1*k_2*k_3*k_4$ multiply-accumulate operations to compute an output tensor. If use a dictionary-based convolution computation, we need
$L_1*L_2*L_3*L_4*|D|$ MACs and not more than $O(|D|*output\_tensor\_size)$ addition operation. Number of «add» operation depends on specific case but it is usually less than the number of multiplications. Taking into account relative simplicity of the «add» operation (much less gates than it is required for «multiply» implementation) we did not include them into analysis. One important thing to take into account is that intermediate memory storage responses to dictionary elements. This memory amount depends on dictionary size but as we see from estimations it is negligibly small.

There is a natural tradeoff between complexity and accuracy, given for every dictionary size $|D|$. Larger dictionary size tends to lower approximation error, consequently to higher accuracy, which is fully matches with tests. To save overall network performance on the highest level, various layers should be decomposed in different ways. Empirically, we found that layers closer to the network entry point harder yields the decomposition and to save the performance a larger dictionary size should be taken. In contrast, layers which are close to network end could be decomposed stronger. This phenomenon and test results are shown in Section 4.

To find the approximation $\widehat{W}$ of kernel tensor $W$ and to form a dictionary we split $W$ along the data propagation directions. This is not a unique way to split the tensor $W$. Our choice of this particular way is explained by the fact that in our experiments we found that best possible slicing method has almost the same dictionary size. The best possible solution (when no neighborhood restriction applied to components of dictionary elements, and we choose remotely situated position to form dictionary element) has extreme complexity and we did not manage to look over all possible treatments. But the approach we used seems to be a good approximation of this best possible one.

## 4. Section 4.

To test presented approach, we used Imagenet classification benchmark and the VGG-16 net. It is also compared with the CP-tensor decomposition [6]. The proposed method outperforms tensor decomposition by a factor of \*\*\*2\*\*\*. Codes for dictionary learning, comparative analysis and tensor decomposition, that we used, can be found \*\*\*here\*\*\*.

The canonic polyadic (CP) tensor decomposition of rank R>0 represents N-dimensional tensor T as a sum of order 1 tensor products.

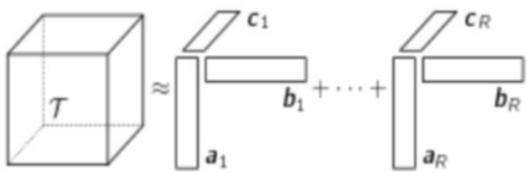

Figure 3. Rank R Canonic Polyadic decomposition of a 3D tensor T.

CP decomposition choice is derived from optimum condition in the Frobenius norm domain.

$$\|T\|_F = \sqrt{\sum_{i,j,k} t_{ijk}^2}, \quad \|T\text{-}CP(T)\|_F \to min$$

CP-tensor decomposition with unknown rank is NP-hard problem [8]. Usually, some tough algorithms like Alternating Least Squares [9] or Tensor Power Method [1] are used for application purposes. During experiments we did not observe a significant difference between results of these methods, so we choose ALS implemented in TensorLab 3.0 [9] as method to find decomposition.

CP-decomposition considers a tensor in special form of «orthogonal» vector product. This form can be used for efficient convolution computation in the way, presented in [6]. First, convolution is calculated with one vector of decomposition, next the second vector is applied, finally the third vector is applied to convolve the result of previous operations. It is not hard to see that results will be the same if to convolve input 1 time with tensor product of 3 vectors. For the N-d case this operation will have N steps.

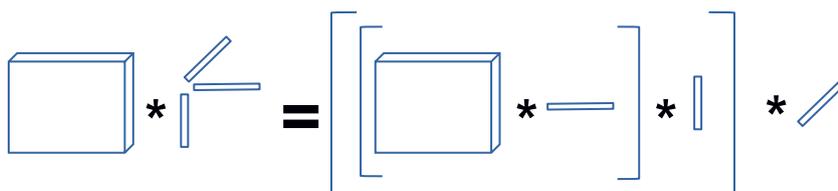

Figure 4. Use CP-decomposition to compute convolution with input tensor in fast way.

Testing methodology was taken from Imagenet benchmark: VGG-16 was used for classification task on ImageNet validation dataset consist of 50000 images. Number of classes to distinguish was equal to 1000. As accuracy metrics we used top-1 and top-5 accuracy. In a Table 1 we show impact of one layer decomposition to overall net accuracy for proposed method and for CP-tensor decomposition. One can see, coarser decomposition tends to lower accuracy.

|  | CFQ 200 centroids | CFQ 500 centroids | CFQ 700 Centroids | Tensor CP rank 400 | Tensor CP rank 700 |
|---|---|---|---|---|---|
| Layer 3_1 | 0.618 | 0.645 | 0.649 | 0.641 | 0.65 |
| Layer 3_2 | 0.626 | 0.645 | 0.649 | 0.629 | 0.646 |
| Layer 3_3 | 0.629 | 0.648 | 0.648 | 0.631 | 0.646 |

Table 1. Impact of CP-tensor decomposition and filter quantization with clustering of layer 3 of VGG-16. Baseline accuracy is 0.65

| Method specification | Layers decomposition (centroid number or rank of CP) | Speed-up ratio | Top-1 accuracy |
|---|---|---|---|
| Baseline VGG-16 |  | 1x | 0.65 |
| Filter quantization + Clustering v.1 | 350,400,800,900,700,600,400, 350,350,300,300,250,200 | 10.9x | 0.642 |
| CP-decomposition v.1 | 13,100,125,225,325,450,450,450, 650,1000,1000,1000,1000,1000 | 2.3x | 0.65 |
| CP-decomposition v.2 | 8,60,75,135,195,270,270,390 600,600, 600,600,600 | 3.8x | 0.642 |

Table 2. Impact of decomposition of all layers of the VGG16. We smallest CP-decomposition ranks and number of centroids to achieve this level of accuracy. Further decreasing of rank or centroid number leads to accuracy loss.

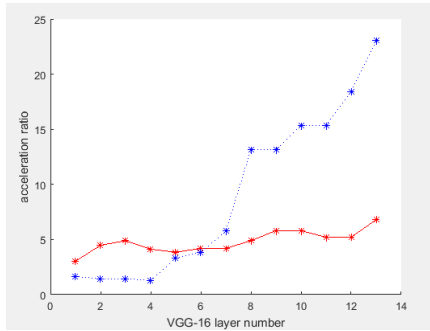

Different layers can be decomposed with different precision. Experiments shows, that the farther layer from net's input, the coarser decomposition could be performed without significant loss of accuracy. This observation is right both for CP-tensor decomposition and filter quantization followed by clustering. Figure 5 shows dependency between maximum possible speed-up ratio (in order not to decrease overall top-1 error more than by 1%) and layer number.

Figure 5. Effect of decomposition speed up per every convolutional layer of VGG-16. Dashed lines devotes to our method, solid devotes to tensor decomposition.

## 5. Section 5.

Here we enclose method to accelerate and compress convolutional layers of neural networks. Method based on convolution quantization followed by clustering. We compared this method with CP-tensor decomposition and found it outperforms it by a large margin both in compression and acceleration. Method does not require net retraining and will be useful for CNN hardware designers as it requires convolution special convolution operation. Method could be used in common with other acceleration/compression approaches like quantization, pruning or knowledge distillation.